%% file: article.tex
\documentclass[conference]{IEEEtran}
\IEEEoverridecommandlockouts
\usepackage[]{cite}
\usepackage{verbatim}
\usepackage{amsmath,amssymb,amsfonts}
\usepackage{algorithmic}
\usepackage{graphicx}
\usepackage{textcomp}
\usepackage{xcolor}

\usepackage{siunitx}
\usepackage{multirow}
\usepackage{cleveref}
\usepackage[caption=false,font=footnotesize]{subfig}

\usepackage{makecell}
\usepackage{tikz}
\usepackage{pgfplots}
\pgfplotsset{compat=1.18}
\usepgfplotslibrary{colorbrewer,groupplots,fillbetween}
\usetikzlibrary{shapes, arrows.meta, positioning, calc, backgrounds, fit, shadows, decorations.markings, 3d, fadings, decorations.pathreplacing}

\definecolor{tabblue}{HTML}{1F77B4}
\definecolor{taborange}{HTML}{FF7F0E}
\definecolor{tabgreen}{HTML}{2CA02C}
\definecolor{tabred}{HTML}{D62728}
\definecolor{tabpurple}{HTML}{9467BD}
\definecolor{tabbrown}{HTML}{8C564B}
\definecolor{tabpink}{HTML}{E377C2}
\definecolor{tabgray}{HTML}{7F7F7F}
\definecolor{tabolive}{HTML}{BCBD22} 
\definecolor{tabcyan}{HTML}{17BECF}

\definecolor{zoneBlue}{RGB}{235, 245, 255}
\definecolor{zoneGreen}{RGB}{235, 250, 235}
\definecolor{zonePurple}{RGB}{245, 235, 250}
\definecolor{zoneRed}{RGB}{255, 240, 240}
\definecolor{zoneOrange}{RGB}{255, 248, 235}
\definecolor{borderGray}{RGB}{180, 180, 180}

\definecolor{plasma0}{RGB}{13, 8, 135}

\definecolor{plasma1}{RGB}{34, 6, 144}

\definecolor{plasma2}{RGB}{49, 5, 151}

\definecolor{plasma3}{RGB}{63, 4, 156}

\definecolor{plasma4}{RGB}{78, 2, 162}

\definecolor{plasma5}{RGB}{91, 1, 165}

\definecolor{plasma6}{RGB}{103, 0, 168}

\definecolor{plasma7}{RGB}{116, 1, 168}

\definecolor{plasma8}{RGB}{129, 4, 167}

\definecolor{plasma9}{RGB}{141, 11, 165}

\definecolor{plasma10}{RGB}{152, 20, 160}

\definecolor{plasma11}{RGB}{162, 29, 154}

\definecolor{plasma12}{RGB}{173, 39, 147}

\definecolor{plasma13}{RGB}{182, 48, 139}

\definecolor{plasma14}{RGB}{191, 57, 132}

\definecolor{plasma15}{RGB}{199, 66, 124}

\definecolor{plasma16}{RGB}{207, 76, 116}

\definecolor{plasma17}{RGB}{214, 85, 109}

\definecolor{plasma18}{RGB}{221, 94, 102}

\definecolor{plasma19}{RGB}{227, 104, 95}

\definecolor{plasma20}{RGB}{233, 114, 87}

\definecolor{plasma21}{RGB}{239, 124, 81}

\definecolor{plasma22}{RGB}{243, 135, 74}

\definecolor{plasma23}{RGB}{247, 145, 67}

\definecolor{plasma24}{RGB}{250, 158, 59}

\definecolor{plasma25}{RGB}{252, 169, 52}

\definecolor{plasma26}{RGB}{253, 181, 46}

\definecolor{plasma27}{RGB}{253, 194, 41}

\definecolor{plasma28}{RGB}{252, 208, 37}

\definecolor{plasma29}{RGB}{249, 221, 37}

\definecolor{plasma30}{RGB}{245, 235, 39}

\definecolor{plasma31}{RGB}{240, 249, 33}

\pgfdeclareverticalshading{plasmabar}{100bp}{
  color(0bp)=(plasma0);
color(3bp)=(plasma1);
color(6bp)=(plasma2);
color(9bp)=(plasma3);
color(12bp)=(plasma4);
color(16bp)=(plasma5);
color(19bp)=(plasma6);
color(22bp)=(plasma7);
color(25bp)=(plasma8);
color(29bp)=(plasma9);
color(32bp)=(plasma10);
color(35bp)=(plasma11);
color(38bp)=(plasma12);
color(41bp)=(plasma13);
color(45bp)=(plasma14);
color(48bp)=(plasma15);
color(51bp)=(plasma16);
color(54bp)=(plasma17);
color(58bp)=(plasma18);
color(61bp)=(plasma19);
color(64bp)=(plasma20);
color(67bp)=(plasma21);
color(70bp)=(plasma22);
color(74bp)=(plasma23);
color(77bp)=(plasma24);
color(80bp)=(plasma25);
color(83bp)=(plasma26);
color(87bp)=(plasma27);
color(90bp)=(plasma28);
color(93bp)=(plasma29);
color(96bp)=(plasma30);
color(100bp)=(plasma31)
}

\newcommand{\plasmacolorbar}[2]{%
\begin{tikzpicture}[x=1pt,y=1pt,baseline=(current bounding box.center)]
  \shade[shading=plasmabar] (0,0) rectangle (#1,#2);
  \draw[line width=0.3pt] (0,0) rectangle (#1,#2);

  \draw[line width=0.3pt] (#1,0) -- (#1+3,0);
  \draw[line width=0.3pt] (#1,0.5*#2) -- (#1+3,0.5*#2);
  \draw[line width=0.3pt] (#1,#2) -- (#1+3,#2);

  \node[anchor=west,font=\footnotesize] at (#1+5,0) {0};
  \node[anchor=west,font=\footnotesize] at (#1+5,0.5*#2) {0.5};
  \node[anchor=west,font=\footnotesize] at (#1+5,#2) {1};
\end{tikzpicture}%
}

\def\BibTeX{{\rm B\kern-.05em{\sc i\kern-.025em b}\kern-.08em
    T\kern-.1667em\lower.7ex\hbox{E}\kern-.125emX}}
\begin{document}

\title{Self-Supervised Representation Learning via Hyperspherical Density Shaping}

\author{\IEEEauthorblockN{Esteban Rodríguez-Betancourt}
\IEEEauthorblockA{\textit{Posgrado en Computación e Informática} \\
\textit{Universidad de Costa Rica}\\
esteban.rodriguezbetancourt@ucr.ac.cr}
\and
\IEEEauthorblockN{Edgar Casasola-Murillo}
\IEEEauthorblockA{\textit{Escuela de Ciencias de la Computación} \\
\textit{Universidad de Costa Rica}\\
edgar.casasola@ucr.ac.cr}
}

\maketitle

\begin{abstract}
Modern self-supervised representation learning methods often relies on empirical heuristics that are not theoretically grounded. In this study we propose HyDeS, a theoretically grounded method based on multi-view mutual information maximization within an hyperspherical space using Shannon differential entropy with a non-parametric von Mises-Fisher density estimator.

We show that HyDeS bias the trained model towards focusing on foreground features of the images and perform well on segmentation tasks such as VOC PASCAL, while it lags in fine-grained classification. We provide a detailed analysis of the induced latent space geometry and learning dynamics, that can be used for designing other theoretically grounded self-supervised learning methods.
\end{abstract}


\input{method_diagram}

\section{Introduction}\label{sec:introduction}
Self-supervised methods keep improving, but much of the progress has come from heuristics: architectural asymmetries, stop-grads, whitening, variance penalties, and carefully tuned augmentations. These choices work, yet the connection to first principles is often indirect. Ideally, self-supervised learning methods should be grounded on information theory. However, to properly design such methods, we must understand how each principle builds toward a good representation and at which point the principle is not enough to guarantee good performance.

In this work, we propose HyDeS, a particular implementation of multi-view mutual information maximization maximization on a hyperspherical space, using Shannon differential entropy with a non-parametric von Mises-Fisher density estimator. Our goal is to test how this theoretical backed formulation behaves, study what it is actually learning and where it breaks compared to modern SLS methods.

We found that our method can learn meaningful representations, however it lags behind in fine-grained separation. Regardless, it shows good dense segmentation, surprising alignment with WordNet and word embeddings similarities on ImageNet-1k and exihibit an emergent object-centric organization with a bias towards ``foreground features''. Additionally, we performed several experiments to determine how certain hyperparameters affect the classification accuracy, and found that weakening the global separation makes accuracy improvement slower, but leads to higher accuracy given enough epochs, while a stronger global separation leads to earlier higher accuracy but eventual learning stagnation.

\section{Related Work}\label{sec:related_work}

One of the earliest theories behind learning was the InfoMax principle \cite{linsker1988infomax}. The InfoMax principle states that the mutual information between the input and the output of an artificial neural network must be maximized, leading to learning of useful features \cite{linksker1988selforganization}. It later grounded classic Independent Component Analysis via entropy maximization at the output layer \cite{Bell1995}. A complementary line formalizes representation learning through explicit mutual information terms and variational bounds \cite{barber2003imalgo,belghazi18a}.

Several deep methods are framed around maximizing mutual information between inputs and learned representations. Deep InfoMax pursues both global and local agreements \cite{hjelm2019dim}. Augmented Multiscale Deep InfoMax (AMDIM) extends this with multi-scale and patch-level objectives \cite{bachman2019}. Contrastive predictive coding casts InfoMax through a classification view and the InfoNCE bound \cite{oord2019representationlearningcontrastivepredictive}, an idea that is later exploited by methods such as SimCLR and MoCo \cite{chen2020simclr,he2020moco}. Alternatively, other methods maximize MI between cluster assignments for segmentation or cluster discovery \cite{ji2019invariant}.

Beyond InfoNCE, used in SimCLR, there are broader MI estimators and f-divergence bounds used in practice \cite{barber2003imalgo,belghazi18a,Nguyen2007}. Work in information-theoretic learning also explores entropy and divergence criteria directly \cite{Principe2010}.

Most modern SSL methods do not rely on InfoMax theoretical grounding. The reason is that strictly maximizing $I(X;Z)$ do not lead necessarily to a good or unique solution \cite{tschannen2020mutualinformationmaximizationrepresentation}. A parallel theoretical foundation argues for shaping information rather than maximizing it: preserve what predicts the task while discarding nuisance variability \cite{tishby2000informationbottleneckmethod}. Many recent SSL objectives can be read as mixing mutual information maximization of other variables with implicit bottlenecks or regularizers, whitening, variance penalties, stop-grad, or architectural asymmetries \cite{zbontar2021barlow,bardes2021vicreg,grill2020bootstrap}.

In our case, we are doing multi-view mutual information maximization, $I(Z_1;Z_2)$. Unlike similar approaches such as InfoNCE, we impose a hyperspherical space and we use density of many views of the same image to calculate the differential entropy.

\section{Method Description}\label{sec:approach}

Our method is based on maximizing mutual information between augmented views. To achieve this, we approximate the mutual information between views $\mathcal{I}(Z_1; Z_2)$ by using an entropy decomposition, where both marginal and conditional entropies are estimated using a kernel density estimator on the hypersphere. A diagram summarizing how our method works is shown in \Cref{fig:method_workflow}.

Let $Z_1$ and $Z_2$ be the continuous representations of two augmented views of the same source. The mutual information across views is defined as $\mathcal{I}(Z_1;Z_2) = \mathcal{H}(Z) - \mathcal{H}(Z_1 | Z_2)$, where $\mathcal{H}(Z)$ is the differential entropy of $Z$ and $\mathcal{H}(Z_1 | Z_2)$ is the conditional differential entropy.

Calculating differential entropy in high-dimensional spaces is usually considered intractable, and the metric can be cheated by the model by pushing dimension magnitudes towards infinity. To simplify calculations and eliminate the unbounded growing  risks, we restrict the representation geometry to a hypersphere, bounding all similarities between $-1$ and $1$ by using cosine similarity. 

Differential entropy is defined as $\mathbb{E}\left[-\log(P(Z))\right]$, where $P$ is the probability density function of the embedding distribution $Z$. As we do not know $P$, we approximate it using a Kernel Density Estimator. Specifically, under the hyperspherical geometry we imposed, we can use the von Mises-Fisher (vMF) kernel to estimate the continuous probability density in a non-parametric way.

Then von Mises-Fisher kernel is defined as follows. For unit vectors $Z_i, Z_j \in \mathbb{S}^{D-1}$ with cosine similarity 
$s_{ij} = Z_i^\top Z_j$, 
the von Mises–Fisher (vMF) kernel with concentration $\kappa$ and normalization constant $C_D(\kappa)$ is:
\begin{equation}
K_\kappa(Z_i,Z_j)
\;=\;
C_D(\kappa)\,\exp\!\big(\kappa\, s_{ij}\big),
\label{eq:vmf-kernel}
\end{equation}

Now, we can estimate the global differential entropy over a mini-batch $\mathcal{B}$ as:

\begin{equation}
    \mathcal{H}_{\mathrm{global}} = \mathbb{E}_{Z_i \sim \mathcal{B}} \left[ -\log \left( \mathbb{E}_{Z_j \sim \mathcal{B}-\{Z_i\}} \left[
    K_\kappa(Z_i,Z_j)
    \right]
    \right) \right]
    \label{eq:global-entropy}
\end{equation}

and the local differential entropy (only between the set $\mathcal{P}(i)$ of positive pairs of $i$), to enforce augmentations invariance, can be expressed as:

\begin{equation}
\mathcal{H}_{\mathrm{local}} = \mathbb{E}_{Z_i \sim \mathcal{B}} \left[ -\log\left( \mathbb{E}_{Z_j \sim \mathcal{P}(i)} \left[ K_\kappa(Z_i,Z_j) \right]\right) \right]
\label{eq:local-entropy}
\end{equation}

Finally, to facilitate experiments, we added two $\alpha$ and $\beta$ factors, however only $\alpha=\beta=1$ matches the proper maximization of mutual information between views. So, we can express the mutual information formula as:
\begin{equation}
\mathcal{I}(Z_1;Z_2) \;\approx\; \alpha \cdot \mathcal{H}_{\mathrm{global}} - \beta \cdot \mathcal{H}_{\mathrm{local}}
\label{eq:mi-objective}
\end{equation}

For generating the view augmentations of each sample, we opted for a similar augmentation recipe as DINO \cite{caron2021dino}. In our case, we used two global views and six local views. Each global view takes a crop of 40\% to 100\% of the sample. And each local view takes a crop of 5\% to 40\% of the sample.

\section{Experimental Methodology}

To rigorously evaluate the properties of the representations learned via de HyDeS objective, we designed an experimental setup to evaluate structural characteristics, semantic alignment and downstream classification accuracy.

We trained and evaluated our method on CIFAR-10/CIFAR-100 \cite{krizhevsky2009learning}, STL-10 \cite{coates2011stl10}, Food-101 \cite{bossard14} and ImageNet-1k \cite{deng2009imagenet} datasets. We used the network architectures ResNet-18 \cite{he2016resnet}, ResNet-50 (for ImageNet-1k) and ViT-Tiny \cite{dosovitskiy2020vit} (for STL-10). The training was done on a single NVIDIA H100 GPU with 80GB of GPU memory hosted in the cloud.

Unless mentioned otherwise, all trainings were done using AdamW optimizer, a batch size of 128 and a learning rate of $10^{-3}$. To make our runs reproducible, we implemented our code using the Lightly Framework and used the evaluation setup developed by \cite{kalapos2024whiteningconsistentlyimprovesselfsupervised}. Models trained on on CIFAR-10, CIFAR-100, STL-10, and Food-101 were trained for 200 epochs. We used the default augmentations settings for each method, and in our case we took the DINO augmentation pipeline with 2 global views and 6 local views.

For ImageNet-1k, we decided to use instead the pretrained checkpoints from Lightly Framework for SimCLR, BYOL, SwAV, Barlow Twins, VICReg and DINO, which were trained for 100 epochs and used a batch size of 256. For ReSA \cite{weng2025resa}, we used their published checkpoint of 100 epochs and batch size 256. Finally, for the supervised baseline and HyDeS, we trained it ourselves, using a batch size of 512.

We evaluated HyDeS in two classification tasks: class classification using linear probe and KNN, and dense segmentation using VOC Pascal \cite{Everingham10pascalvoc}. Then, we analyze the induced latent space geometry, its semantic alignment and PCA visualization of learned features. For the geometric analysis, we measured anisotropy, feature correlation, center vector norm, centroid rank, global embedding rank, sensitivity index, mean pairwise angle and sparsity. For showing the emergent semantic alignment we compared the inter-class distances with the ones provided by WordNet and a textual embedding model (all-MiniLM-L2-v2).

\section{Downstream Task Results}
\subsection{Classification Results}\label{subsec:ClassificationResults}

The classification accuracy is summarized in \Cref{tab:accuracy-table}. While the results were good for datasets with a low number of ``latent classes'', it is clear that our method underperformed in datasets with harder to distinguish classes, such as ImageNet-1k and Food-101.

\begin{table}[tb]
    \centering
\caption{Final Linear and K-NN accuracy (top-1 and top-5)}
    \label{tab:accuracy-table}
\begin{tabular}{lllcccc}
    & & &
    \multicolumn{2}{c}{\bfseries Linear} & 
    \multicolumn{2}{c}{\bfseries K-NN} \\
    \multicolumn{1}{c}{\bfseries } & 
    \multicolumn{1}{c}{\bfseries Method} & 
    \multicolumn{1}{c}{\bfseries Arch} &
    \multicolumn{1}{c}{\bfseries Top 1} & \multicolumn{1}{c}{\bfseries Top 5} &
    \multicolumn{1}{c}{\bfseries Top 1} & \multicolumn{1}{c}{\bfseries Top 5} \\
    \hline
\multirow{8}{*}{\rotatebox[origin=c]{90}{\bfseries STL-10}}
    & \textcolor{gray}{Supervised} & \textcolor{gray}{RN18} & \textcolor{gray}{71.86} & \textcolor{gray}{97.41} & \textcolor{gray}{72.43} & \textcolor{gray}{87.69} \\
    & SimCLR & RN18 & \textbf{82.89} & \textbf{99.36} & \textbf{79.16} & \textbf{94.91} \\
    & BYOL & RN18 & 81.28 & 99.28 & 77.14 & 94.39 \\
    & BT & RN18 & 82.19 & 99.04 & 78.20 & 93.98 \\
    & VICReg & RN18 & 81.69 & 99.14 & 78.48 & 93.85 \\
    & DINO & RN18 & 79.75 & 99.11 & 76.85 & 93.53 \\
    & HyDeS & RN18 & 81.98 & 99.25 & 76.85 & 93.83 \\
    & HyDeS & VIT-T16 & 85.98 & 99.66 & 80.31 & 94.65 \\
    \hline
\multirow{7}{*}{\rotatebox[origin=c]{90}{\bfseries CIFAR-10}}
    & \textcolor{gray}{Supervised} & \textcolor{gray}{RN18} & \textcolor{gray}{84.17} & \textcolor{gray}{98.97} & \textcolor{gray}{84.16} & \textcolor{gray}{93.36} \\
    & SimCLR & RN18 & 72.58 & 97.86 & 69.16 & 90.25 \\
    & BYOL & RN18 & 69.37 & 97.81 & 65.54 & 89.74 \\
    & BT & RN18 & 74.63 & 97.91 & 71.81 & \textbf{90.87} \\
    & VICReg & RN18 & \textbf{74.72} & \textbf{98.14} & \textbf{72.67} & 90.59 \\
    & DINO & RN18 & 72.46 & 98.02 & 70.07 & 90.34 \\
    & HyDeS & RN18 & 72.54 & 97.97 & 68.11 & 89.14 \\
    \hline
\multirow{7}{*}{\rotatebox[origin=c]{90}{\bfseries CIFAR-100}}
    & \textcolor{gray}{Supervised}        & \textcolor{gray}{RN18} & \textcolor{gray}{53.05} & \textcolor{gray}{78.09} & \textcolor{gray}{51.19} & \textcolor{gray}{69.00} \\
    & SimCLR      & RN18 & 39.93 & 68.86 & 32.92 & 52.92 \\
    & BYOL        & RN18 & 38.68 & 68.84 & 33.91 & 53.72 \\
    & BT          & RN18 & 41.52 & 71.20 & 36.46 & 56.06 \\
    & VICReg      & RN18 & \textbf{43.31} & \textbf{72.38} & \textbf{37.55} & \textbf{57.61}  \\
    & DINO        & RN18 & 39.27 & 68.42 & 34.07 & 53.06 \\
    & HyDeS & RN18 & 38.46 & 69.31 & 32.08 & 53.38 \\
    \hline
\multirow{7}{*}{\rotatebox[origin=c]{90}{\bfseries Food-101}}
    & \textcolor{gray}{Supervised}  & \textcolor{gray}{RN18} & \textcolor{gray}{71.33} & \textcolor{gray}{89.59} & \textcolor{gray}{70.01} & \textcolor{gray}{90.35} \\
    & SimCLR      & RN18 & 61.15 & 85.07 & 49.99 & 77.45 \\
    & BYOL        & RN18 & 58.21 & 82.75 & 45.83 & 73.85 \\
    & BT          & RN18 & 64.11 & 86.66 & 54.90 & 80.15 \\
    & VICReg      & RN18 & \textbf{65.69} & \textbf{87.34} & \textbf{56.61} & 81.19 \\
    & DINO        & RN18 & 64.48 & 87.15 & 55.43 & \textbf{81.60} \\
    & HyDeS & RN18 & 53.69 & 80.10 & 41.55 & 71.05 \\
    \hline
\multirow{9}{*}{\rotatebox[origin=c]{90}{\bfseries \;ImageNet-1k\;}}
    & \textcolor{gray}{Supervised}        & \textcolor{gray}{RN50} & \textcolor{gray}{70.02} & \textcolor{gray}{89.34} & \textcolor{gray}{61.72} & \textcolor{gray}{85.29} \\
    & SimCLR      & RN50      & 62.03 & 84.75 & 45.60 & 74.18 \\
    & BYOL        & RN50      & 61.84 & 84.64 & 45.88 & 74.61 \\
    & SwAV        & RN50      & \textbf{66.15} & \textbf{87.84} & 50.19 & 78.93 \\
    & BT          & RN50      & 59.69 & 82.68 & 46.24 & 74.44 \\
    & VICReg      & RN50      & 62.66 & 84.93 & 47.14 & 75.71 \\
    & DINO        & RN50      & 64.43 & 85.85 & 51.28 & 79.93 \\
    & ReSA        & RN50      & 60.85 & 83.91 & \textbf{60.79} & \textbf{85.58} \\
    & HyDeS & RN50 & 56.62 & 81.34 & 36.64 & 65.65 \\
    \hline
    \rotatebox[origin=c]{90}{\bfseries \;IN-100\;} & \emph{HyDeS*} & RN50 & 68.15 & 89.68 & 55.25 & 77.38
\end{tabular}

* ImageNet-100 was evaluated using the backbone trained on ImageNet-1k.
\end{table}

\subsection{PASCAL Visual Object Classes Segmentation}
We evaluated the trained models on class segmentation, using the PASCAL VOC dataset \cite{Everingham10pascalvoc}. For this evaluation, we took the hypercolumns (the concatenation of the outputs of each layer) of a frozen ImageNet-1k trained ResNet-50 as backbone, and trained a linear probe on top of those representations for 20 epochs using AdamW with a learning rate of $1^{-3}$ and batch size 32. The results are summarized in \Cref{tab:voc-accuracy}. HyDeS had the third highest mIoU of the evaluated models.

\begin{table}[thbp]
    \centering
\caption{PASCAL VOC segmentation via linear probing of hypercolumns of frozen ResNet-50 pretrained on ImageNet-1k}
    \label{tab:voc-accuracy}
\begin{tabular}{lll}
\textbf{Method} & \textbf{Pixel Accuracy (\%)} & \textbf{mIoU (\%)} \\\hline
\textcolor{gray}{Supervised} & \textcolor{gray}{86.34} & \textcolor{gray}{44.40} \\
SimCLR & 86.96 & 45.78 \\
BYOL & 87.27 & 47.98 \\
SwAV & 89.26 & 54.18 \\
Barlow Twins & 86.46 & 44.52 \\
VICReg & 87.58 & 48.97 \\
DINO & \textbf{89.63} & \textbf{56.54} \\
ReSA & 85.92 & 41.40 \\
HyDeS & 87.95 & 49.12 \\

\end{tabular}
\end{table}

Our results show that HyDeS representations contains useful information to perform segmentation, evidenced by the higher pixel accuracy and mean intersection over union (mIoU). Additionally, this shows an interesting tradeoff, where geometries with similar accuracy using a linear probe behave significantly different over another task. This evidences that it is not appropriate to evaluate representation quality by classification accuracy alone.

\section{Learned Representation Analysis}

\subsection{Emergent object-centric organization}
Learned representations could be potentially encoding any part of the image. To determine what aspects are being encoded by HyDeS we used SmoothGrad-CAM on a ResNet-50 trained on ImageNet-1k and attention map for the ViT-Tiny trained on STL-10.

The SmoothGrad-CAM visualization in \Cref{fig:smoothgrad-overlay} shows that the model tend to focus on foreground features ignoring the background. For instance, it focuses more on the biplane, pelicans, cats and the base of the statue. However, it didn't focused much on the horse, instead prefering to focus on the sea waves. In the ViT-Tiny attention map at \Cref{fig:attention-overlay}, the model is much more focused on foreground features, however its attention dilutes a bit into background features. Both visualizations show that our method induces a bias towards foreground features of the image.

\newcommand{\smoothgradsample}[1]{\includegraphics[width=0.24\linewidth]{images/#1.jpg} \includegraphics[width=0.24\linewidth]{smoothgrad/#1_smoothgrad.jpg}}

\begin{figure}[tb]
    \centering
    \smoothgradsample{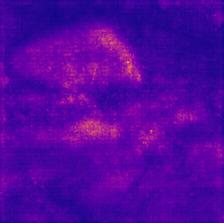}
    \smoothgradsample{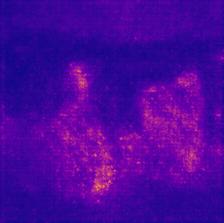}\\\vspace{1mm}
    \smoothgradsample{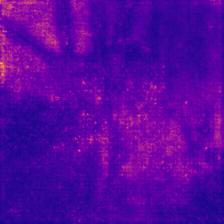}
    \smoothgradsample{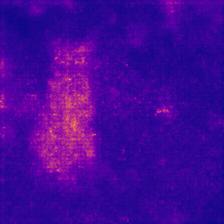}\\\vspace{1mm}
    \smoothgradsample{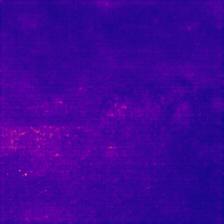}
    \smoothgradsample{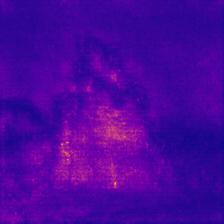}
    \caption{SmoothGrad-CAM visualization of gradients produced by a ResNet-50 trained on ImageNet-1k, across all layers.}
    \label{fig:smoothgrad-overlay}
\end{figure}

\newcommand{\attentionsample}[1]{\includegraphics[width=0.24\linewidth]{images/#1.jpg} \includegraphics[width=0.24\linewidth]{attention/#1_attn.jpg}}

\begin{figure}[tb]
    \centering
    \attentionsample{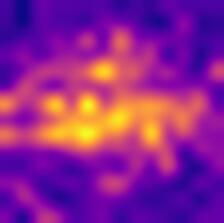}
    \attentionsample{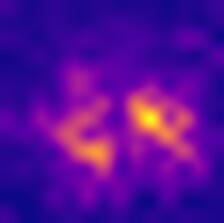}\\\vspace{1mm}
    \attentionsample{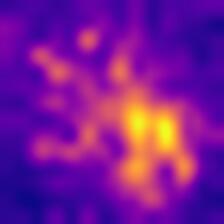}
    \attentionsample{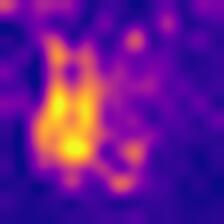}\\\vspace{1mm}
    \attentionsample{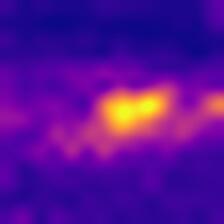}
    \attentionsample{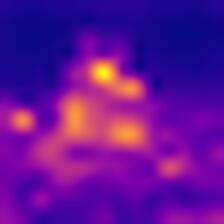}
    \caption{Attention map from a ViT-Tiny trained on STL-10, obtained by averaging CLS-to-patch attention across all transformer blocks.}
    \label{fig:attention-overlay}
\end{figure}

The gradients or attention heatmaps of the previous subsection can give us an idea of where the model is putting its attention. However, it does not mean that the model learned to identify whether the sections of the image belong to the same object or not. To investigate that, we took the outputs of the internal layers, scaled them, concatenated the embeddings and then produced a PCA projection of the three most important components into RGB space. The resulting visualizations can be seen in \Cref{fig:segmentations}. The PCA visualization shows that the model learned to associate similar representations to regions that belong to the same ``object''. For instance, the airplane, pelicans, cats and horses tend to be represented with the same color (per image). Additionally, background features such as grass, sand or the sky is also represented consistently across the same image. These differentiation of parts of the image may explain the comparatively good results on VOC Pascal dense segmentation.

\newcommand{\segmentationsample}[1]{\includegraphics[width=0.24\linewidth]{images/#1.jpg} \includegraphics[width=0.24\linewidth]{segmentation/#1_pca_rgb_multilayer.jpg}}

\begin{figure}[tb]
    \centering
    \segmentationsample{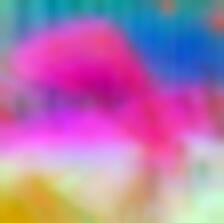}
    \segmentationsample{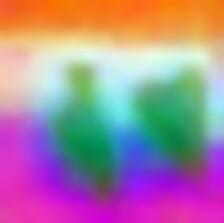}
    \segmentationsample{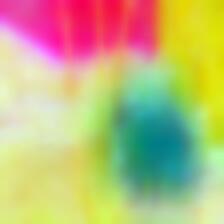}
    \segmentationsample{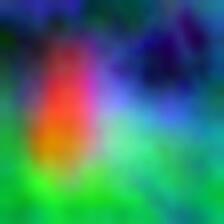}
    \segmentationsample{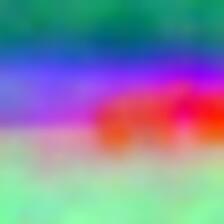}
    \segmentationsample{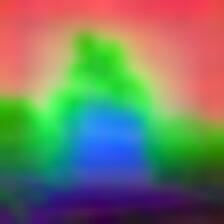}
    \caption{PCA visualization of hypercolumns of ResNet-50 trained on ImageNet-1k.}
    \label{fig:segmentations}
\end{figure}

\subsection{Geometry Analysis}\label{subsec:GeometryAnalysis}

Different learning methods induces vastly different latent space geometries. We compare properties of the latent spaces of the evaluated methods in two datasets: ImageNet-1k which has 1000 classes and Food-101 that has 101 classes. In ImageNet-1k there is a mix of easily separable classes and harder classes (dog breeds, for instance). Food-101 separation is much harder. \Cref{tab:qualityMetrics} summarizes several  measurements of the latent space geometry induced by each evaluated method.

\begin{table*}[htbp]
\caption{Geometric Properties of Representations}
    \label{tab:qualityMetrics}
\begin{center}
    \begin{small}
      \begin{sc}
\begin{tabular}{
clcccccccc
}
\hline
& Method & Aniso. & Corr. & CVN & \makecell{Centroid\\Rank} & \makecell{Embed.\\Rank} & $d^\prime$ & \makecell{All Pairwise\\Angle (Mean/Std)} & \makecell{Positive Pairwise\\Angle (Mean/Std)} \\
    \hline
\multirow{9}{*}{\rotatebox[origin=c]{90}{\bfseries ImageNet-1k}}
& \textcolor{gray}{Supervised} & \textcolor{gray}{0.29} & \textcolor{gray}{\textbf{0.04}} & \textcolor{gray}{0.53} & \textcolor{gray}{498} & \textcolor{gray}{\textbf{1662}} & \textcolor{gray}{1.70} & \textcolor{gray}{$73.62^\circ \pm 3.72^\circ$} & \textcolor{gray}{$64.22^\circ \pm 6.95^\circ$} \\
& SimCLR     & 0.26 & 0.06 & 0.52 & 412 & 1390 & 1.71 & $74.42^\circ \pm 5.48^\circ$ & $61.25^\circ \pm 9.98^\circ$\\
& BYOL       & 0.35 & 0.08 & 0.69 & 364 & 1227 & 1.67 & $60.99^\circ \pm 7.28^\circ$ & $46.22^\circ \pm 10.71^\circ$\\
& SwAV       & 0.24 & 0.05 & 0.51 & 454 & 1459 & 1.82 & $74.56^\circ \pm 4.93^\circ$ & $58.59^\circ \pm 12.13^\circ$\\
& Barlow Twins & 0.39 & 0.07 & 0.63 & 353 & 1312 & 1.71 & $66.40^\circ \pm 5.50^\circ$ & $53.71^\circ \pm 9.33^\circ$\\
& VICReg     & 0.25 & 0.06 & 0.51 & 455 & 1390 & 1.81 & $74.85^\circ \pm 5.30^\circ$ & $58.44^\circ \pm 12.27^\circ$\\
& DINO       & 0.53 & 0.08 & 0.74 & 304 & 963 & \textbf{2.03} & $56.16^\circ \pm 5.14^\circ$ & $\mathbf{40.97^\circ} \pm 9.89^\circ$\\
& ReSA       & 0.21 & \textbf{0.04} & 0.46 & \textbf{532} & 1563 & 2.00 & $77.60^\circ \pm 4.20^\circ$ & $62.61^\circ \pm 10.44^\circ$\\
& HyDeS & \textbf{0.12} & 0.07 & \textbf{0.37} & 316 & 1103 & 1.96 & $\mathbf{82.17^\circ} \pm 8.61^\circ$ & $55.99^\circ \pm 18.24^\circ$\\
    \hline
\multirow{7}{*}{\rotatebox[origin=c]{90}{\bfseries Food-101}}
& \textcolor{gray}{Supervised} & \textcolor{gray}{0.49} & \textcolor{gray}{0.075} & \textcolor{gray}{0.69} & \textcolor{gray}{53} & \textcolor{gray}{375} & \textcolor{gray}{\textbf{1.41}} & \textcolor{gray}{$61.27^\circ \pm 5.25^\circ$} & \textcolor{gray}{$53.23^\circ \pm 6.13^\circ$}\\
& SimCLR       & 0.41 & 0.087 & 0.65 & 46 & 332 & 1.05 & $64.57^\circ \pm 7.54^\circ$ & $55.93^\circ \pm 9.21^\circ$\\
& BYOL         & 0.70 & 0.109 & 0.84 & 26 & 251 & 0.80 & $44.52^\circ \pm 5.27^\circ$ & $\mathbf{40.23}^\circ \pm 5.56^\circ$\\
& Barlow Twins           & \textbf{0.21} & 0.077 & \textbf{0.48} & \textbf{64} & 365 & 1.27 & $\mathbf{76.45^\circ} \pm 7.87^\circ$ & $63.74^\circ \pm 12.70^\circ$\\
& VICReg       & 0.28 & 0.078 & 0.54 & 61 & 360 & 1.26 & $72.54^\circ \pm 7.82^\circ$ & $60.57^\circ \pm 11.40^\circ$\\
& DINO         & 0.28 & \textbf{0.064} & 0.54 & 56 & \textbf{394} & 1.23 & $72.77^\circ \pm 5.40^\circ$ & $64.49^\circ \pm 8.36^\circ$\\
& HyDeS        & 0.32 & 0.141 & 0.58 & 37 & 260 & 1.01 &  $69.76^\circ \pm 10.03^\circ$ & $58.52^\circ \pm 13.09^\circ$\\
    \hline
\end{tabular}
\end{sc}
\end{small}
\end{center}
\end{table*}

\begin{figure*}[thbp]
    \centering

    \newlength{\panelw}
    \setlength{\panelw}{0.18\textwidth}

    \subfloat[Supervised]{
        \includegraphics[width=\panelw]{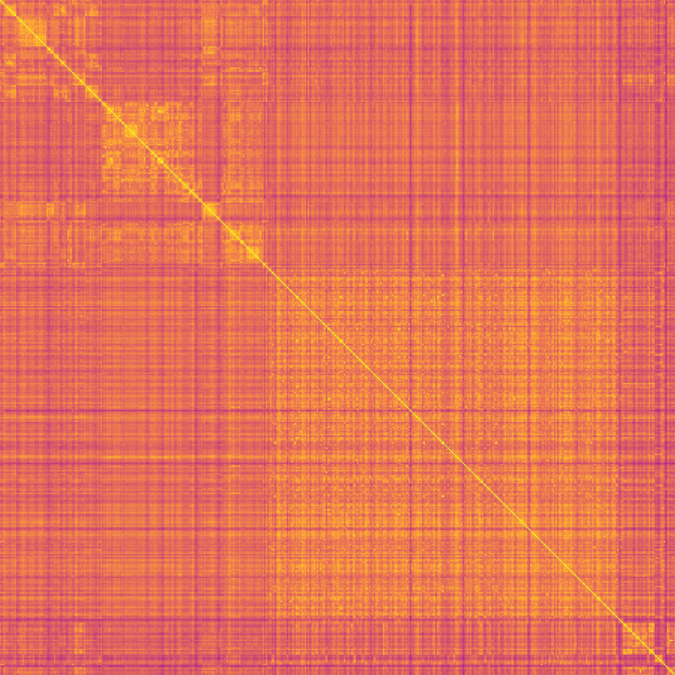}
    }\hfill
    \subfloat[SimCLR]{
        \includegraphics[width=\panelw]{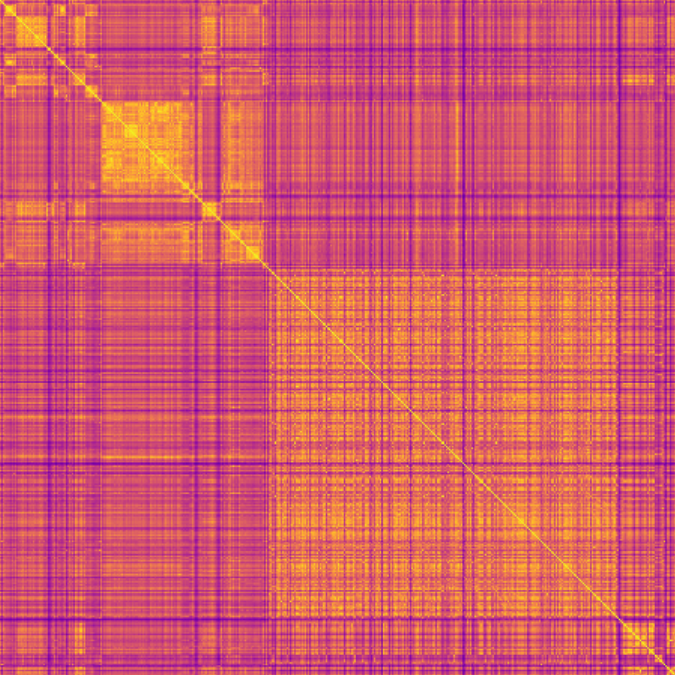}
    }\hfill
    \subfloat[BYOL]{
        \includegraphics[width=\panelw]{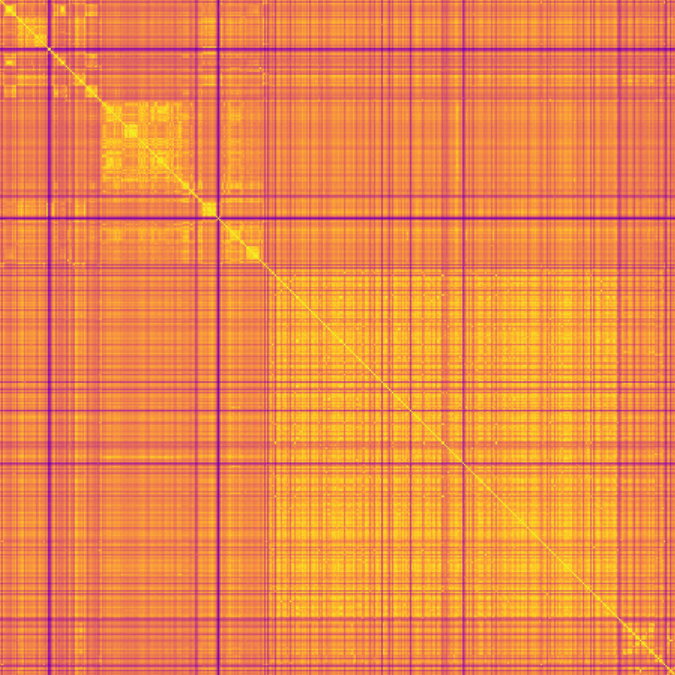}
    }\hfill
    \subfloat[SwAV]{
        \includegraphics[width=\panelw]{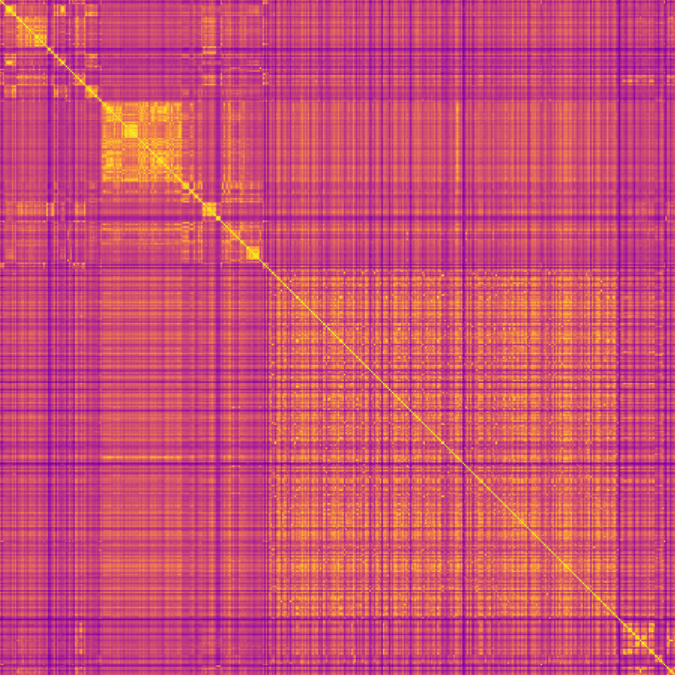}
    }\hfill
    \subfloat[Barlow Twins]{
        \includegraphics[width=\panelw]{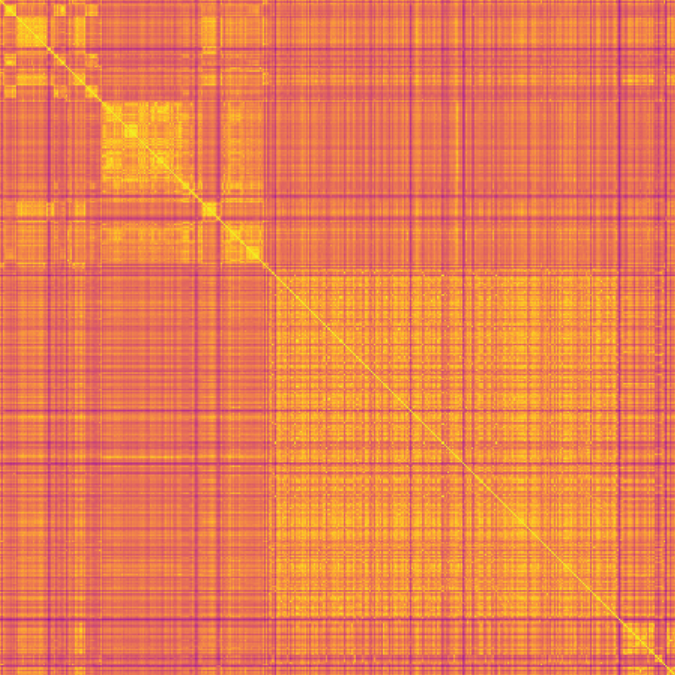}
    }

    \vspace{0.6em}

    \makebox[\textwidth][c]{%
        \subfloat[DINO]{
            \includegraphics[width=\panelw]{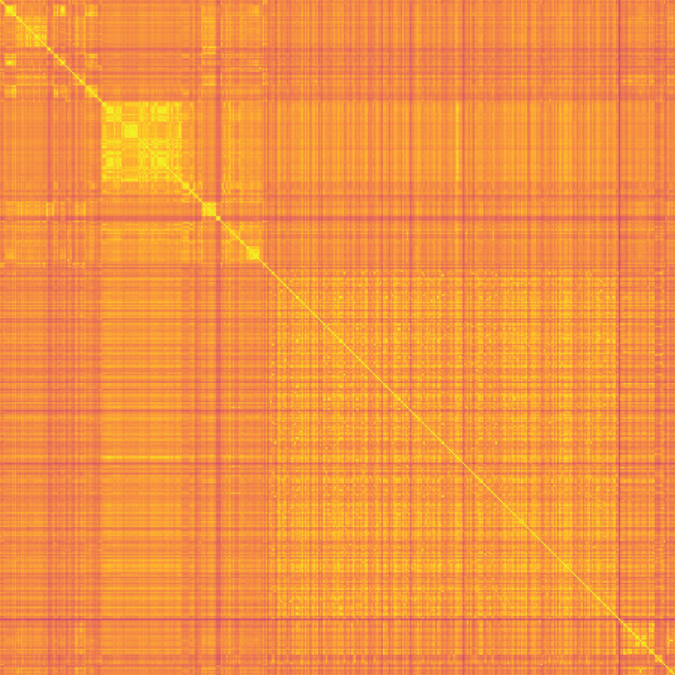}
        }\hspace{0.02\textwidth}
        \subfloat[VICReg]{
            \includegraphics[width=\panelw]{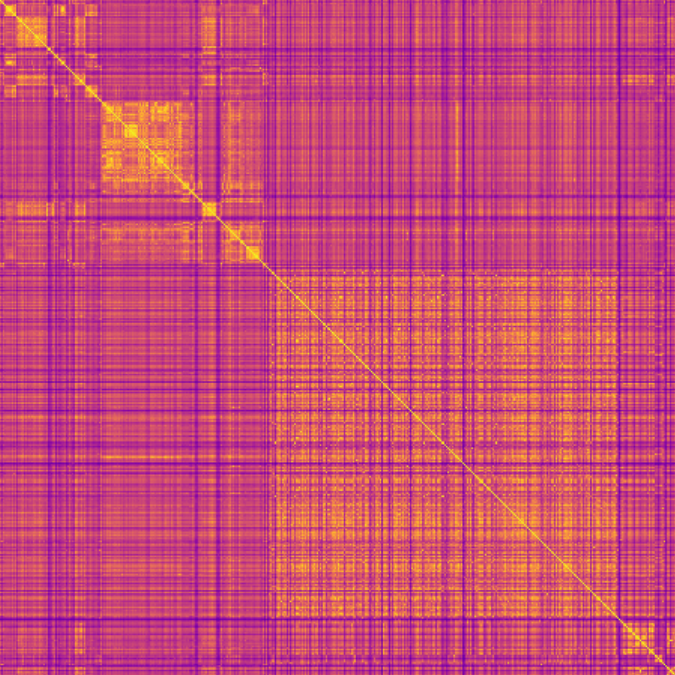}
        }\hspace{0.02\textwidth}
        \subfloat[ReSA]{
            \includegraphics[width=\panelw]{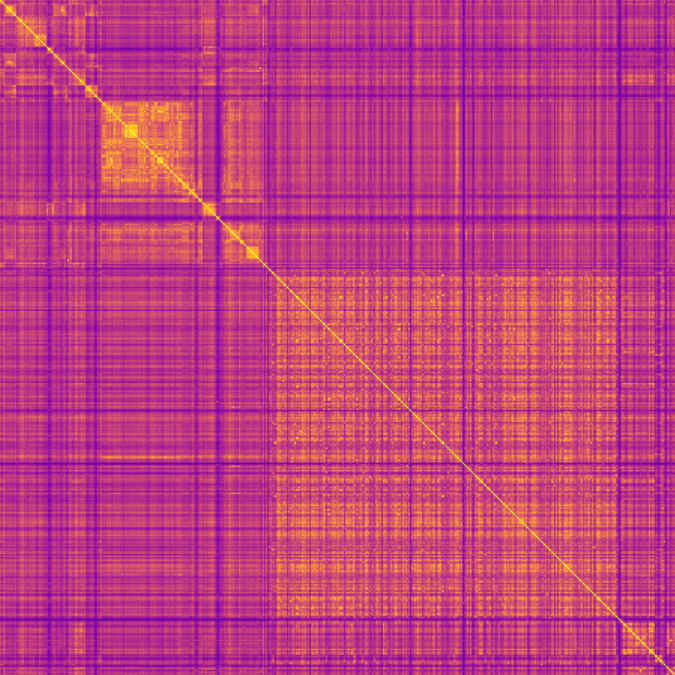}
        }\hspace{0.02\textwidth}
        \subfloat[HyDeS ($\kappa=1$)]{
            \includegraphics[width=\panelw]{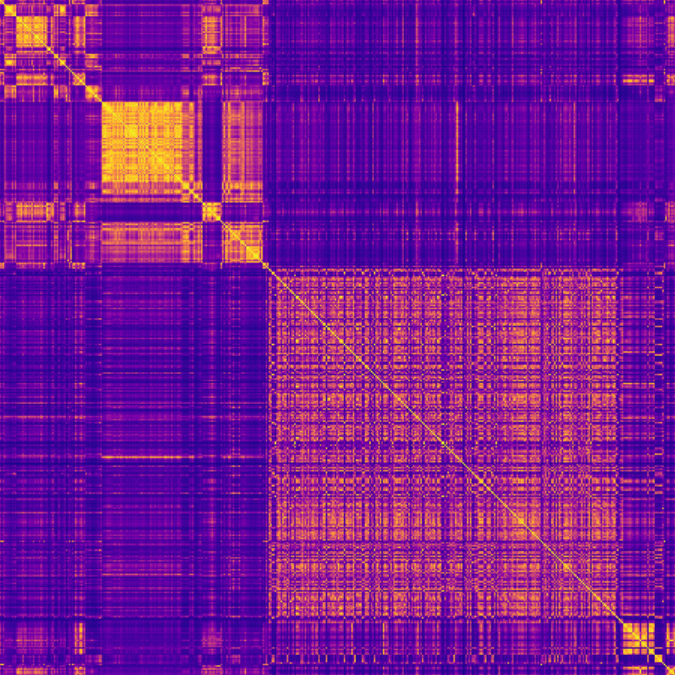}
        }\hspace{0.02\textwidth}
        \makebox[0.2\panelw][c]{%
            \raisebox{0.5\panelw}[0pt][0pt]{%
                \plasmacolorbar{12}{100}%
            }%
        }
    }

    \caption{Pairwise cosine similarity between ImageNet-1k class centroids. Brighter means more similarity, while darker colors mean bigger angles (0 is equivalent to 90º).}
    \label{fig:centroid_pairwise_similarity_all}
\end{figure*}

In datasets that do not require strong fine grain separation, HyDeS achieves the lowest anisotropy, lowest center vector norm \cite{jha2024commonstabilitymechanismselfsupervised} and highest mean pairwise angle. A lower center vector norm should lead to a better distributed latent space. Under the uniformity and alignment framework \cite{wangisola2020}, HyDeS achieved better uniformity, however the alignment (positive pairwise angle) is worse than other methods.

A consequence of having in average higher angles between classes is that class separation seems to emerge from a ``near-orthogonal'' representation, rather than a dense description of learned features. This can visualized in \Cref{fig:centroid_pairwise_similarity_all}, that shows the pairwise cosine similarity between the class centroids of ImageNet-1k classes. HyDeS representation is much darker (closer to $90^\circ$), while the other methods mean angle is much less aggressive, signaling that their learned features are shared between much more classes. Interestingly, these two representation strategies can lead to similarly high sensitivity $d'$, where a higher value indicates better class discrimination.

The same HyDeS plot in \Cref{fig:centroid_pairwise_similarity_all} signals that HyDeS struggles with fine-grained separation: the bright yellow block correspond to dog breeds, which are not separated as well as the other classes. This failure mode is more evident in a hard dataset such as Food-101, where HyDeS metrics are overall worse.

Effective Rank is a metric where HyDeS is consistently worse compared to other evaluated methods. This may be caused due to the global density minimization objective being too strong and breaking semantic structure. This conclusion is supported by the $\beta$ ablation, described in \Cref{subsec:LocalEntropyWeight}.

\subsection{Emergent Semantic Alignment}\label{subsec:SemanticAlignment}
Learned representations may encode similarity between entities or classes. Similarly represented classes may also signal that the model confuses them. In order to determine how well the learned representations encode similarity we measured the spearman correlation between relative distances in the representations with distances in WordNet and text embeddings. For WordNet, we used Wu-Palmer Similarity (WUP) and Leacock-Chodorow Similarity (LCH), which measures how similar two word senses are, based on the shortest path that connects the senses in WordNet. Additionally, we also measured the cophenetic correlation, between each learned representation and WordNet when using agglomerative clustering. Finally, we measured Spearman correlation with the distances of a text embedding model, specifically MiniLM-L6-v2. The results are summarized in \Cref{tab:semanticAlignment}.

\begin{table}[htbp]
    \centering
    \caption{Spearman Correlation of Relative Distances between WordNet and Learned Representations on ImageNet (p-value $< 0.05$)}
    \label{tab:semanticAlignment}
    \begin{tabular}{ccccc}
\hline
Method       & WuP & LCH & Cophenetic & MiniLM \\ \hline
Supervised   & 0.29 & 0.31 & 0.47 & 0.08 \\
SimCLR       & 0.35 & 0.38 & 0.26 & 0.13 \\
BYOL         & 0.22 & 0.28 & 0.18 & 0.02 \\
SwAV         & 0.25 & 0.28 & 0.19 & 0.06 \\
Barlow Twins & 0.31 & 0.35 & 0.21 & 0.11 \\
VICReg       & 0.25 & 0.28 & 0.18 & 0.09 \\
DINO         & 0.25 & 0.27 & 0.17 & 0.08 \\
ReSA         & 0.30 & 0.33 & 0.24 & 0.10 \\
HyDeS        & 0.50 & 0.51 & 0.65 & 0.19 \\
\hline
    \end{tabular}
\end{table}

\section{Impact of Hyperparameters}\label{sec:Hyperparameters}
We explored variations of the number of dimensions in the projector, the impact of different $\kappa$ (bandwidth parameter) in the final accuracy and the weight of the local entropy estimation ($\beta$ parameter). Unlike other methods such as Barlow Twins, we did not find any noticeable difference for changing the number of dimensions of the projector. On the other hand, $\kappa$ and $\beta$ influence is more noticeable.

\subsection{Projector Dimensions}
\begin{figure}[htbp]
    \centering
    \includegraphics[width=\linewidth]{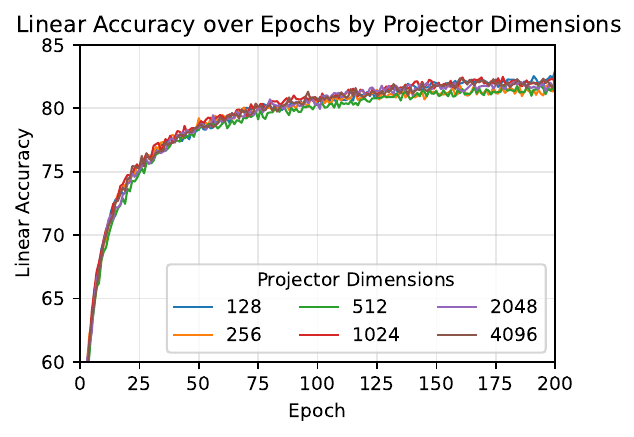}
    \caption{Linear probe top 1 accuracy per epoch, varying the number of dimensions of the projector.}
    \label{fig:dimensions-plot}
\end{figure}

We trained a ResNet-18 using our method on STL-10, changing the number of dimensions of the final projector to 128, 256, 512, 1024, 2048 and 4096. There is no clear advantage of increasing the size of the final projector, and the learning curves are very similar in all cases, as shown in \Cref{fig:dimensions-plot}.

\subsection{Bandwidth Parameter}\label{subsec:BandwidthParameter}

\begin{figure}[htbp]
    \centering
    \includegraphics[width=\linewidth]{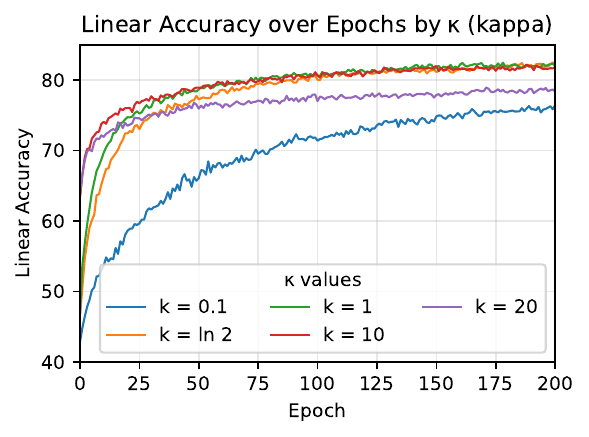}
    \caption{Linear probe top 1 accuracy per epoch, varying the bandwidth parameter.}
    \label{fig:kappa-plot}
\end{figure}

We trained a ResNet-18 using our method on STL-10, varying the bandwidth parameter $\kappa$ to $0.1$, $0.6931 \approx \ln 2$, $1$, $10$ and $20$. Recall that $\kappa$ is the concentration (bandwidth) of the vMF kernel used in our non-parametric density estimate: larger $\kappa$ produces a sharper kernel with small angular spread, and a smaller $\kappa$ a smoother one with a larger angular spread. In this case, we found that higher $\kappa$, such as $\kappa=20$, values accelerate early accuracy, however, if it is too high it hinders the rest of the learning process after the initial epochs. Smaller $\kappa$ values lead to slower accuracy improvements, but given enough epochs they eventually catch up and surpass models trained with higher $\kappa$. In the \Cref{fig:kappa-plot} the accuracy per epoch is shown for each $\kappa$ evaluated.

\subsection{Local Entropy Estimation Weight}\label{subsec:LocalEntropyWeight}
\begin{figure}[tb]
    \centering
    \includegraphics[width=\linewidth]{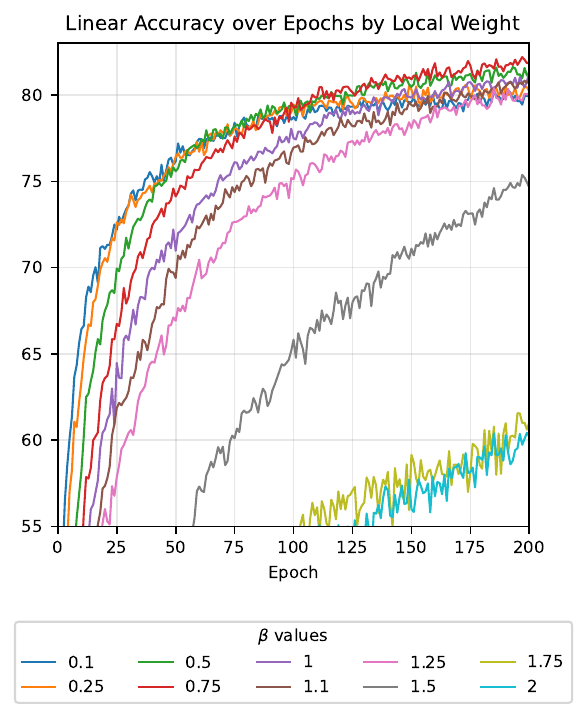}
    \includegraphics[width=\linewidth]{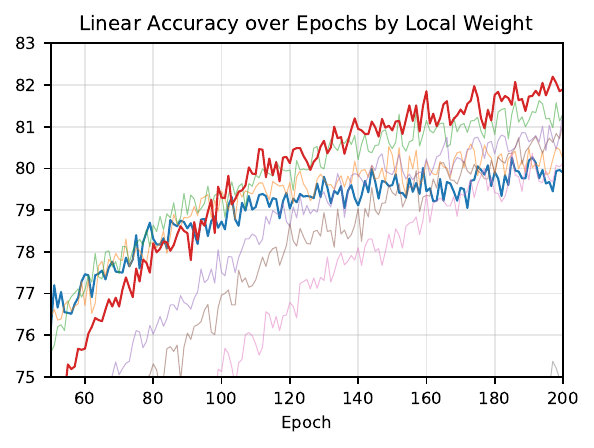}
    \caption{Linear accuracy over epochs by changing the $\beta$ weight (above). Below, a zoom in, highlighting the curves for $\beta=0.1$ (red, final accuracy higher) and $\beta=0.75$ (blue, final accuracy lower).}
    \label{fig:beta-and-zoom}
\end{figure}

Additionally, we modified the weight of the local entropy estimation. We found that reducing the $\beta$ weight, alternatively making separation heavier, leads to higher early accuracy. However, these setups seems to get stuck after some epochs, and runs with higher local weight end up having a higher final accuracy, despite the slower early accuracy gains. The accuracy during training of those variations is shown in \Cref{fig:beta-and-zoom}. Below in that figure, a zoom in section is shown, where we can appreciate how the accuracy of $\beta=0.1$ stalls compared to other settings, particularly $\beta=0.75$. This suggests that longer training would benefit from a softer separation force. Which is interesting, as this suggests that at least for this sort of classification problems, the best representation may not be the one with maximum mutual information.

\section{Discussion}
Our work shows that our multi-view mutual information maximization method is able to learn good representations and the method tends to learn much faster than other methods at least in the early epochs. However, unfortunately these early gains typically are accompanied with a overall lower quality in late stages of the training. Ideally, we would like to learn fast and also get better even representations as we train the model further.

In our analysis, it is evident that what may be preventing the network from learning better representations may be the geometry itself. Due to our geometry and loss objective, the expansion loss term of our method is trying to push away each point as much as possible. This condition may be too strong for learning good representations, as we found that weakening the expansion leads to better accuracy results. So, learning may benefit from early separation, but beyond some point this organization is not useful for further learning. This contrasts with other entropy maximization methods such as SimCLR, which instead uses a weighting mecanism to decide how far to push away negatives. If framed under the general objectives of contrastive learning presented in \cite{wangisola2020}, our alignment term should be similar to most other methods, but our expansion term tries to maximize separation as much as possible, leading to early accuracy gains but late learning stalling.

For future work, we consider important to analyze what made this model to be a faster early learner, and what caused it to start lagging compared to other methods at late training epochs. Studying this aspect may lead to methods strongly grounded on information theory, that learn faster and produce even better representations.

\section{Conclusion}\label{conclusion}

In this paper, we discussed an implementation of multi-view mutual information maximization. In this case, we opted to use hypersphere geometry, non-parametric kernel density estimators and differential Shannon entropy. We found that this method can generate good representations in datasets with low number of classes, but it starts to lag behind for datasets with higher level of classes. However, the metrics suggest that lower accuracy is due to a learned coarse structure, and not a failure to learn good representations. Our experiments additionally point in the direction that we can get better representations if we relax our global entropy term. This suggests that, at least for classification, representations should not look to maximize mutual information. Instead there must be another theoretical objective that should explain better how to produce strong fine-grained structures.

As future work, we are exploring self-supervised learning techniques that preserve quick early learning and still maintain competitive quality over longer training schedules.

\bibliographystyle{ieeetr}
\bibliography{references}

\end{document}

%% file: method_diagram.tex
\begin{figure*}[thbp!]
\begin{center}
\begin{tikzpicture}[
    font=\sffamily\footnotesize,
    >={Latex[width=2mm,length=2mm]},
    block/.style={
        rectangle, draw=black!70, fill=white, rounded corners=2pt, 
        minimum height=0.7cm, align=center, drop shadow={opacity=0.15}, inner sep=3pt
    },
    loss/.style={
        rectangle, draw=black!80, fill=red!10, rounded corners=4pt, 
        minimum height=0.6cm, font=\sffamily\footnotesize\bfseries, inner sep=3pt
    },
    vector/.style={
        rectangle, draw=black!80, fill=white, minimum width=0.6cm, minimum height=0.6cm, 
        font=\bfseries\footnotesize, align=center
    },
    imgnode/.style={inner sep=0pt, drop shadow={opacity=0.2}, draw=white, line width=1pt},
    sum/.style={circle, draw=black, fill=white, inner sep=2pt},
    arrow/.style={->, thick, draw=black!70},
    dashed_arrow/.style={->, dashed, thick, draw=black!60},
    bus/.style={thick, draw=black!60, rounded corners=4pt},
    stopgrad/.style={
        decoration={markings, mark=at position 0.6 with {\draw[red, thick] (-2pt,-2pt) -- (2pt,2pt);\draw[red, thick] (-2pt,2pt) -- (2pt,-2pt);}}, 
        postaction={decorate}
    },
    zone/.style={rounded corners, draw=borderGray, inner sep=0pt}
]

    \node[imgnode, label={[font=\sffamily\footnotesize\bfseries]below:Input $X$}] (input) at (0,0) {
        \includegraphics[width=1.6cm]{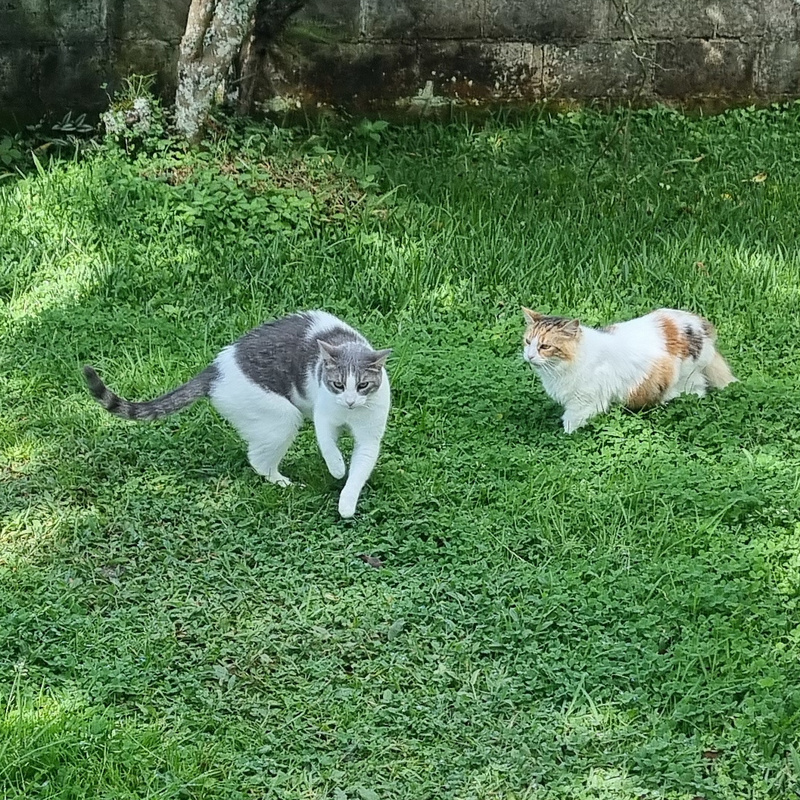}
    };

    \node[imgnode] (aug1) at (2.0, 1.4) {
        \includegraphics[width=0.9cm, trim={150 200 100 100}, clip]{images/cat5.jpg}
    };

    \node[imgnode] (aug2) at (2.0, 0) {
        \includegraphics[width=0.9cm, trim={100 200 400 300}, clip]{images/cat5.jpg}
    };
    
    \node[imgnode] (aug3) at (2.0, -1.4) {
        \includegraphics[width=0.9cm, trim={450 300 20 150}, clip]{images/cat5.jpg}
    };

    \draw[arrow] (input.east) -- (aug1.west);
    \draw[arrow] (input.east) -- (aug2.west);
    \draw[arrow] (input.east) -- (aug3.west);

    \node[block, fill=green!10, text width=1.4cm, minimum height=4.6cm] (encoder) at (3.8, 0) {Shared\\Encoder\\$f_\theta$};
    
    \node[block, fill=blue!5, text width=1.0cm, minimum height=4.6cm] (l2norm) at (5.6, 0) {Proj.\\$\mathbb{S}^{D-1}$};

    \draw[arrow] (aug1.east) -- (encoder.west |- aug1.east);
    \draw[arrow] (aug2.east) -- (encoder.west |- aug2.east);
    \draw[arrow] (aug3.east) -- (encoder.west |- aug3.east);
    
    \draw[arrow] (encoder.east |- aug1.east) -- (l2norm.west |- aug1.east);
    \draw[arrow] (encoder.east |- aug2.east) -- (l2norm.west |- aug2.east);
    \draw[arrow] (encoder.east |- aug3.east) -- (l2norm.west |- aug3.east);

    \node[vector, anchor=west] (z1) at ([xshift=0.2cm, yshift=1.4cm]l2norm.east) {$Z_1$};
    \node[vector, anchor=west] (z2) at ([xshift=0.2cm]l2norm.east) {$Z_2$};
    \node[vector, anchor=west] (z3) at ([xshift=0.2cm, yshift=-1.4cm]l2norm.east) {$Z_3$};

    \draw[thick, black!70] (l2norm.east |- aug1.east) -- (z1.west);
    \draw[thick, black!70] (l2norm.east |- aug2.east) -- (z2.west);
    \draw[thick, black!70] (l2norm.east |- aug3.east) -- (z3.west);

    \coordinate (LossLeft) at (9.25, 0);  
    \coordinate (LossRight) at (13.5, 0);
    
    \coordinate (LocalY) at (0, 1.4);
    \node[block, fill=violet!5, text width=2.4cm] (vmf_local) at (LossLeft |- LocalY) {vMF Density\\$\mathbb{E}_{\mathcal{P}(i)} [K_\kappa(Z_i, Z_j)]$};
    \node[block, fill=violet!5, right=0.4cm of vmf_local, text width=2.2cm] (log_local) {Exp. Neg-Log\\$\mathbb{E}_{\mathcal{B}} [-\log(\cdot)]$};

    \draw[arrow] (vmf_local) -- (log_local);

    \coordinate (GlobalY) at (0, -1.4);
    \node[block, fill=red!5, text width=2.4cm] (vmf_global) at (LossLeft |- GlobalY) {vMF Density\\$\mathbb{E}_{\mathcal{B}} [K_\kappa(Z_i, Z_j)]$};
    \node[block, fill=red!5, right=0.4cm of vmf_global, text width=2.2cm] (log_global) {Exp. Neg-Log\\$\mathbb{E}_{\mathcal{B}} [-\log(\cdot)]$};
    
    \draw[arrow] (vmf_global) -- (log_global);

    \coordinate (BusMerge) at (7.35, 0);
    
    \draw[bus] (z1.east) -- (BusMerge |- z1.east) -- (BusMerge |- z3.east) -- (z3.east);
    \draw[bus] (z2.east) -- (BusMerge |- z2.east);
    
    \draw[arrow] (BusMerge |- z1.east) -- (BusMerge |- LocalY) -- (vmf_local.west);
    
    \draw[arrow] (BusMerge |- z3.east) -- (BusMerge |- GlobalY) -- (vmf_global.west);

    \node[sum] (sigma) at (14, 0) {$\sum$};
    \node[right=0.5cm of sigma, font=\sffamily\footnotesize\bfseries] (total) {$\mathcal{I}(Z_1; Z_2)$};

    \draw[arrow] (log_local.east) -| node[pos=0.25, above, font=\footnotesize\bfseries] {$-\beta$} (sigma.north);
    \draw[arrow] (log_global.east) -| node[pos=0.25, below, font=\footnotesize\bfseries] {$\alpha$} (sigma.south);
    \draw[arrow] (sigma) -- (total);

    \begin{pgfonlayer}{background}
        \node[rounded corners, draw=borderGray, fill=zoneBlue, fit=(input)(aug1)(aug3), inner sep=6pt, 
              label={[anchor=north west, inner sep=3pt, font=\sffamily\footnotesize\bfseries]north west:Augmentation}] {};
        
        \path (encoder.north) ++(0, 10pt) coordinate (EncTop);
        \node[rounded corners, draw=borderGray, fill=zoneGreen, inner sep=6pt, fit=(encoder)(z1)(z3)(l2norm)(EncTop), 
              label={[anchor=north, inner sep=3pt, font=\sffamily\footnotesize\bfseries]north:Shared Encoder \& Projection}] {};
        
        \draw[rounded corners, draw=borderGray, fill=zonePurple] 
             (7.7, 0.6) rectangle (13.6, 2.3);
        \node[font=\sffamily\footnotesize\bfseries, anchor=north west] at (7.75, 2.3) {Local Entropy ($\mathcal{H}_{\text{local}}$)};
        
        \draw[rounded corners, draw=borderGray, fill=zoneRed] 
             (7.7, -0.3) rectangle (13.6, -2.3);
        \node[font=\sffamily\footnotesize\bfseries, anchor=north west] at (7.75, -0.3) {Global Entropy ($\mathcal{H}_{\text{global}}$)};
             
    \end{pgfonlayer}

\end{tikzpicture}
\caption{\textbf{Hyperspherical Density Shaping workflow.} An input image is augmented into multiple views, encoded, and explicitly projected onto the hypersphere $\mathbb{S}^{D-1}$. \textbf{Top (Purple):} The local differential entropy ($\mathcal{H}_{\text{local}}$) is minimized by estimating the vMF density across positive pairs $\mathcal{P}(i)$ to enforce view-invariance. \textbf{Bottom (Red):} The global differential entropy ($\mathcal{H}_{\text{global}}$) is maximized by estimating the vMF density across the entire batch $\mathcal{B}$ to prevent origin collapse and enforce deep space expansion. The final objective balances these forces using hyperparameters $\alpha$ and $\beta$.}
\label{fig:method_workflow}
\end{center}
\end{figure*}